\documentclass[sigconf]{acmart}

\usepackage{amsmath,amsfonts}
\usepackage{algorithmic}
\usepackage{array}
\usepackage[caption=false,font=normalsize,labelfont=sf,textfont=sf]{subfig}
\usepackage{textcomp}
\usepackage{stfloats}
\usepackage{url}
\usepackage{verbatim}
\usepackage{graphicx}
\def\BibTeX{{\rm B\kern-.05em{\sc i\kern-.025em b}\kern-.08em
    T\kern-.1667em\lower.7ex\hbox{E}\kern-.125emX}}
\usepackage{balance}

\usepackage{soul}
\usepackage[utf8]{inputenc}
\usepackage{xcolor}
\usepackage{booktabs}
\usepackage{multirow}
\usepackage{xparse,mleftright}
\usepackage{marvosym}
\usepackage{CJKutf8}
\usepackage{color}
\definecolor{BurntOrange}{rgb}{0.8, 0.33, 0}
\definecolor{OliveGreen}{RGB}{85, 107, 47}
\definecolor{OliveGreen}{RGB}{85, 107, 47}
\definecolor{Plum}{RGB}{139, 102, 139}

\AtBeginDocument{%
  \providecommand\BibTeX{{%
    Bib\TeX}}}

\copyrightyear{2026}
\acmYear{2026}
\setcopyright{cc}
\setcctype{by}
\acmConference[MM '26]{Proceedings of the 34th ACM International Conference on Multimedia}{November 10--14, 2026}{Rio de Janeiro, Brazil.}
\acmBooktitle{Proceedings of the 34th ACM International Conference on Multimedia (MM '26), November 10--14, 2026, Rio de Janeiro, Brazil}
\acmISBN{979-8-4007-2213-4/2026/11}
\acmDOI{10.1145/XXXXXX.XXXXXX}
\begin{document}

\title{Logographic Character Visual Pretraining via Semantic-based Contrastive Learning}


\author{Daqian Shi}
\affiliation{%
  \institution{Queen Mary University of London}
  \city{London}
  \country{UK}}
\email{d.shi@qmul.ac.uk}

\author{Wei Cao}
\affiliation{%
  \institution{Jilin University}
  \city{Changchun}
  \country{China}}
\email{caowei23@mails.jlu.edu.cn}

\author{Xiaoyu Zheng}
\affiliation{%
  \institution{King's College London}
  \city{London}
  \country{UK}}
\email{xiaoyu.zheng@kcl.ac.uk}

\author{Lida Shi}
\affiliation{%
  \institution{Jilin University}
  \city{Changchun}
  \country{China}}
\email{shild21@mails.jlu.edu.cn}

\author{Xiaolei Diao} \thanks{Corresponding author: Xiaolei Diao}
\affiliation{%
  \institution{University College London}
  \city{London}
  \country{UK}}
\email{xiaolei.diao@ucl.ac.uk}

\author{Cédric M. John}
\affiliation{%
  \institution{Queen Mary University of London}
  \city{London}
  \country{UK}}
\email{cedric.john@qmul.ac.uk}

\renewcommand{\shortauthors}{Shi et al.}

\begin{abstract}
Deep learning-based character vision studies, e.g., text recognition, character image denoising, and historical text completion, are offering new solutions for learning, managing, and utilizing character resources. However, the performance of these studies peaks with large and balanced datasets, which is a rarity with real-world character datasets, especially for logographic character languages, e.g., Chinese. The imbalance in data distribution of logographic characters is an issue due to differences in character usage frequency and new characters being continuously created. In this paper, we propose a novel method for logographic character recognition, which introduces a multi-modal learning approach using visual semantics and contextual semantics of characters. A novel pre-training strategy is designed to enhance deep visual representations, especially for datasets suffering from issues of imbalanced and rare instances, by extracting the contextual semantics of each character from the corresponding language models. We conduct experiments across various datasets to evaluate our character recognition method and validate the contrastive pre-training strategy through several downstream tasks. Experimental results demonstrate the superiority of our method compared to state-of-the-art methods.
\end{abstract}



\maketitle

\section{Introduction}
\label{sec:Introduction}

Digitalization has advanced linguistic studies, providing new solutions for learning, managing, and utilizing character resources. Deep representation learning-based character studies have attracted significant attention and serve as a crucial foundation for the development of multimodal text-based models, benefiting tasks such as text understanding and ancient character deciphering \cite{memon2020handwritten}. However, these studies, primarily designed for alphabetic languages, achieve optimal performance only with large and balanced datasets \cite{zhang2017online}, which is a rarity with real-world character datasets, especially for logographic languages. Due to differences in character usage frequency, imbalanced data distribution is a common issue in logographic characters, particularly in historical scenarios \cite{diao2023rzcr}. Moreover, in low-resource languages such as oracle bone script, previously unknown characters continue to emerge through archaeological discoveries, posing inevitable challenges for the effective encoding of such characters \cite{shi2022rcrn}.

\begin{figure}[!t]
\centering
\includegraphics[width=0.99\linewidth]{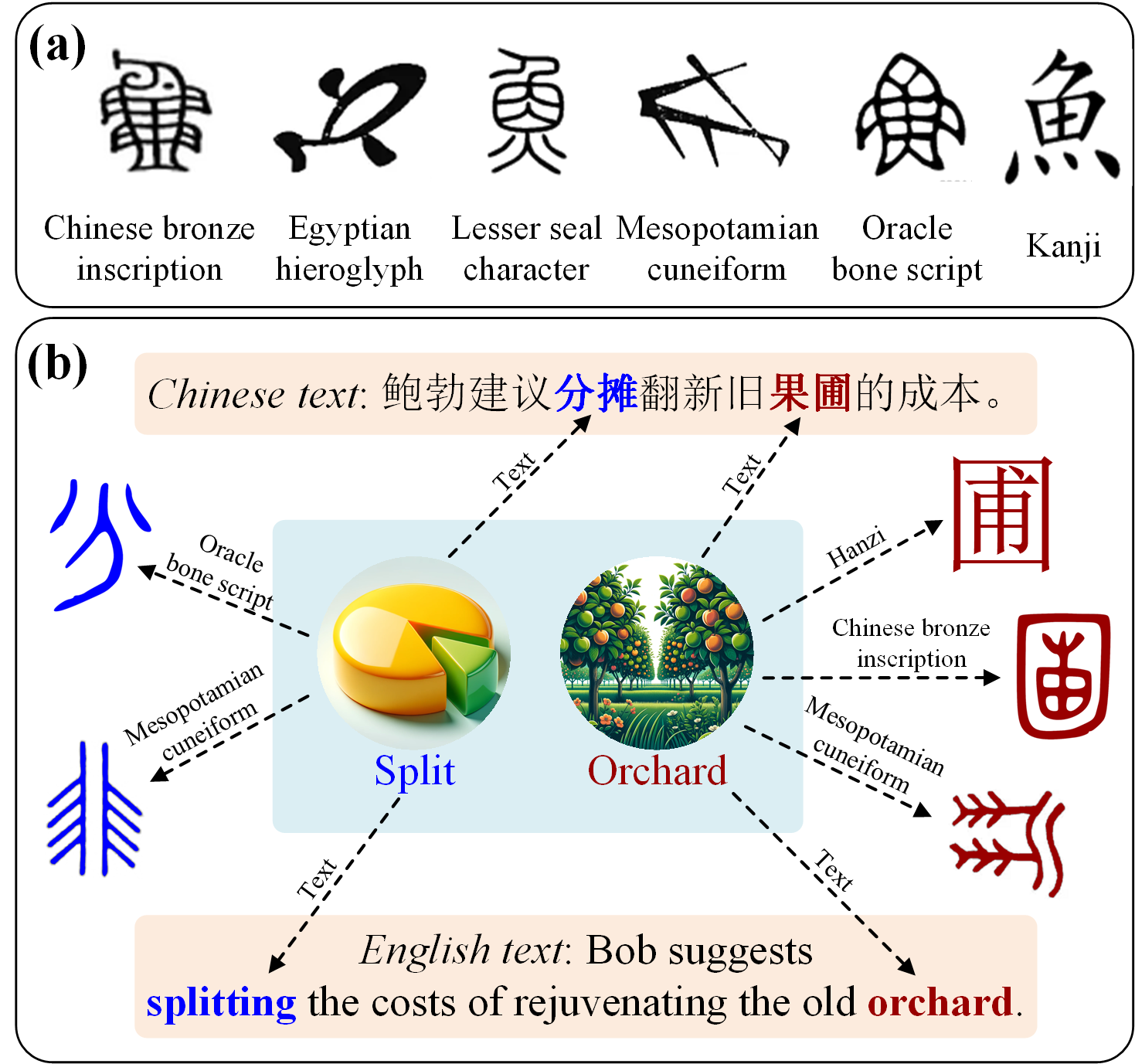} 
\caption{Demonstration of logographic character images and corresponding analysis. (a) the concept of ``fish" represented by different kinds of logographic characters; (b) examples of the consistency between the visual semantics and contextual semantics of each character, where the same concept is highlighted with the same color.}
\label{fig:1}
\end{figure}

Logographic characters refer to characters in a writing system where each character represents a word or a morpheme, e.g., Hanzi in Chinese, Kanji in Japanese, and ancient Egyptian hieroglyphs \cite{zhu2014different}. We observed that such characters were initially designed to present specific semantics through their visual information \cite{hanavan2005hemispheric}, where a specific glyph of each character is applied to abstractly convey its pixel-level features, namely visual semantics. Figure.~\ref{fig:1}(a) shows logographic characters from different languages that express the same concept of ``fish". Abstract fish shapes are obvious in the glyphs, e.g., Egyptian hieroglyphs or Oracle bone scripts. In alphabetic languages, specific semantics can be captured through the context, termed contextual semantics, which promoted the development of recent language models, e.g., GLoVe \cite{pennington2014glove}, and Bert models \cite{kenton2019bert}.  Contextual semantics can also be captured via character embedding when processing logographic languages \cite{nguyen2019hierarchical}. As a result, we observe that the visual and contextual semantics are consistent in terms of the concept each character expresses. Figure.~\ref{fig:1}(b) shows some examples of such observation, where the same concept can be represented by the contextual texts from various languages, and also by various visual shapes of logographic characters. For instance, the concept expressed by the context of the English word ``orchard” is consistent with the Hanzi character ``\begin{CJK}{UTF8}{gbsn}圃\end{CJK}”. 
These intuitions inspire us to improve the current logographic character representation by considering both contextual and visual semantics.

Current logographic character recognition methods evolve from image classification and therefore learn from the correspondence of pixel-level features with the labels of character images \cite{li2023trocr}. For instance, \cite{zhang2017online} combined normalization-cooperated direction-decomposed features with CNN for character recognition, and \cite{nguyen2020semantic} proposed a segmentation-based handwritten Kanji character recognition method using U-Net. Such methods are limited when applied to the real world, as there is a lack of balanced training categories with sufficient samples for real world logographic character sets. Thus, contrastive learning algorithms, e.g., SimCLR \cite{chen2020simple}, have been applied to logographic character recognition \cite{jiang2023group}. The aim is to leverage effective pre-trained models to reduce the impact of imbalanced categories during fine-tuning \cite{xu2019multiple}. 
However, these methods, by simply using a fixed distance (i.e., 1) as the learning target for different categories when learning negative samples, will limit the performance and learning efficiency of character representations \cite{yeh2022decoupled}. Meanwhile, the methods mentioned above rely on learning pixel-level features while neglecting the influence of contextual semantics on the visual representation of logographic characters.

To address the above problems, we propose a novel pre-training strategy that introduces contextual and visual semantics to improve the logographic character representations, especially for datasets suffering from imbalanced data distribution. We introduce language models that extract the contextual semantics of each character, and calculate the latent distance between characters. This semantic relevance, which encodes the relative semantic relationships among characters at the concept level, is employed as a soft label in a contrastive learning framework to constrain the relative positioning of negative samples in the embedding space. We conduct experiments across various datasets with different logographic languages to evaluate our logographic character representation and further validate our contextual-aware pre-training strategy by several downstream tasks. The main contribution of this paper includes:


\begin{itemize}
     \item We propose a novel pretraining strategy aiming to enhance the performance and learning efficiency of imbalanced character categories on their deep representations by introducing semantic-based contrastive learning.
    \item We propose a novel framework that aligns visual and contextual semantics at the distribution level by modeling relative semantic relationships, enabling more effective representation learning for logographic characters.
    \item We conduct experiments across various datasets to evaluate our pre-trained deep representation. Compared with state-of-the-art methods, the experimental results demonstrate the superiority of our method.
\end{itemize}

\section{Related Work}
\label{sec:Related Work}

\subsection{Character Recognition.}

Character recognition has been an area of intensive research that has evolved over the years from rule-based algorithms to deep learning-based models. 
Character recognition primarily revolves around two distinct paradigms: alpha-based character recognition and logographic character recognition, each presenting unique challenges and methodologies. 
Alpha-based character recognition deals mainly with alphabetic languages such as Latin, Cyrillic, or Greek. These languages are characterized by a limited set of characters (typically less than 100), and the recognition process often involves segmenting individual characters followed by classification \cite{cristea2022applying, tallapragada2022greek}. In contrast, logographic character recognition focuses on scripts such as Chinese, Japanese Kanji, or Korean Hanja, where each character represents a word or a morpheme. These scripts feature thousands of unique characters, making recognition more complex \cite{chi2022zinet}. Logographic recognition not only demands the identification of a vastly larger set of characters, but also requires the model to capture more intricate and subtle stroke details and character structures \cite{diao2023toward}. Researchers employ advanced deep learning architectures, including CNNs and attention mechanisms, to effectively handle the complexity and variety of logographic scripts \cite{cirecsan2015multi, zhang2018radical, shi2022charformer}. However, due to the large number and complexity of characters, the recognition of logographic characters is usually limited by an imbalanced data distribution \cite{wang2019radical}. 
Contrastive learning algorithms for pre-training can help reduce the impact of imbalanced categories. This has been demonstrated in representative studies such as using contrastive loss to constrain attention to different sub-regions of characters to enhance the learning of local features \cite{xu2019multiple}, and employing contrastive learning and clustering to encode the visual features of similar characters, thus reducing the dependence on balanced training data \cite{jiang2023group}. 

\subsection{Contrastive Learning.}
Contrastive learning has emerged as a powerful paradigm for self-supervised representation learning. The core idea is to construct positive and negative pairs, learning to make similar instances closer in the feature space while pushing dissimilar instances apart, thereby learning meaningful feature representations.
Early work on contrastive loss function \cite{hadsell2006dimensionality} laid the foundation for contrastive learning. Subsequently, Momentum Contrast (MoCo) \cite{he2020momentum} and SimClr \cite{chen2020simple} further improved the performance of contrastive learning in visual tasks. 
Zhu et al. \cite{zhu2022balanced} proposed class-averaging and class-complement strategies in contrastive learning to address sample imbalance, solving a key challenge in long-tail data distributions. 
%
Recent advances in contrastive learning often focus on modifications in the design of loss functions, data augmentation strategies, and sampling methods to improve representation quality and model generalization \cite{hu2024comprehensive}. However, these methods typically use a fixed distance (typically, 1) as the learning target for negative samples of different classes, which can limit the performance and learning efficiency of the representations.

\section{Intuitive Discussion}
\label{sec:Intuitive Discussion}

\subsection{Character Semantics}

\begin{figure}[!t]
\centering
\includegraphics[width=1\linewidth]{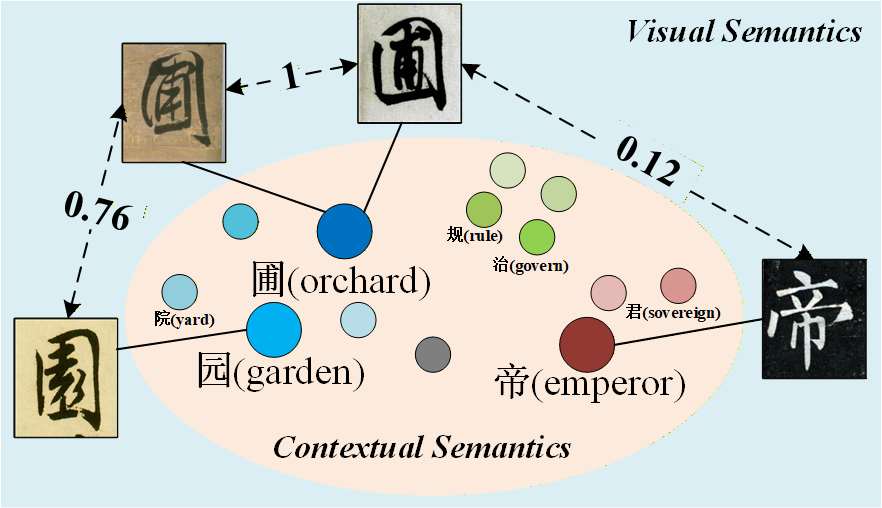}
\caption{Demonstration of the relative semantic neighborhood consistency between contextual semantics and visual semantics of logographic characters, where characters (concepts) with higher relevance tend to be embedded closer (higher similarity index) in both contextual and visual semantic spaces. }
\label{fig:2}
\end{figure}

\begin{figure*}[!t]
\centering
\includegraphics[width=0.9\linewidth]{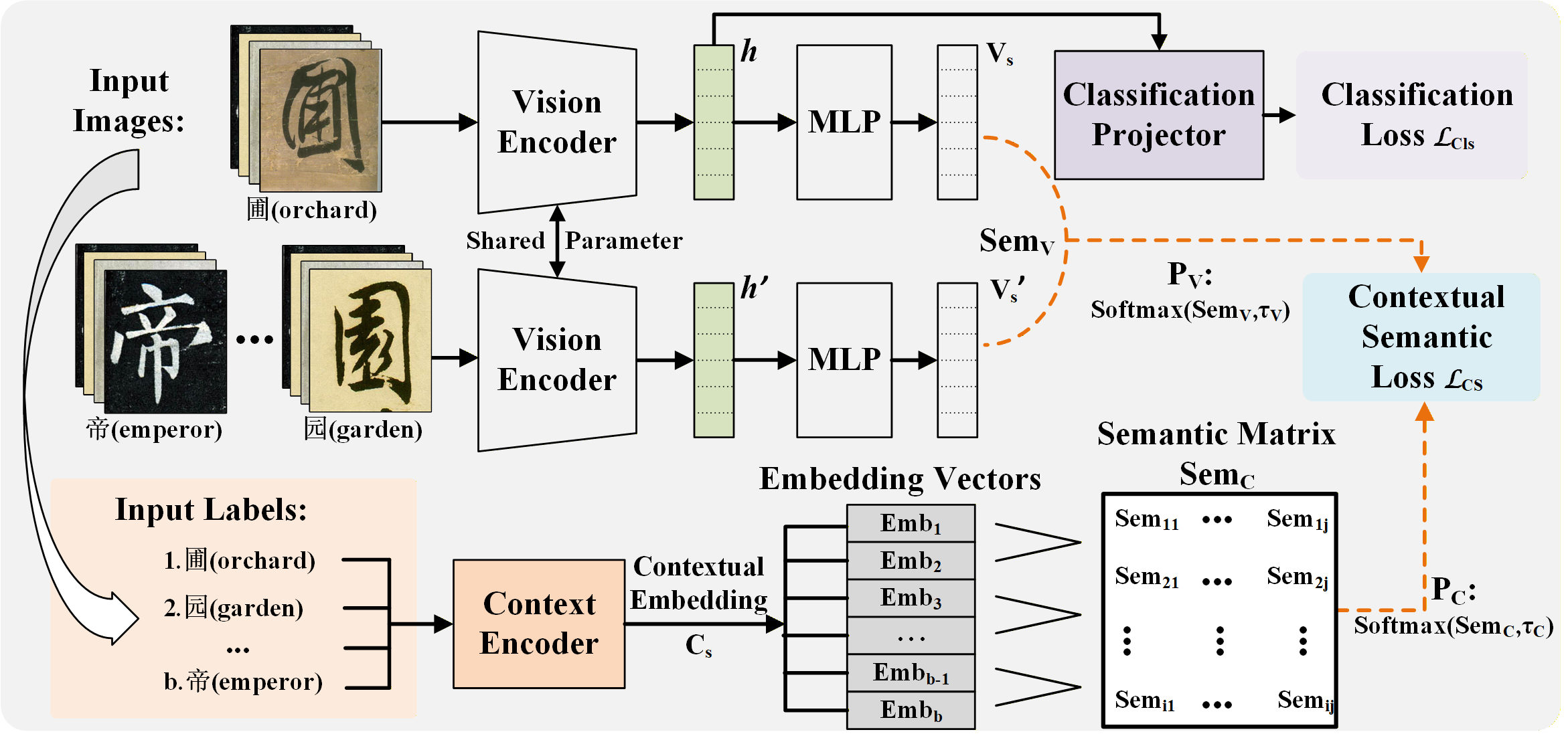}
\caption{The overall architecture of our proposed Logographic Character Recognition method, where different training modules are distinguished by colors. (\textcolor{BurntOrange}{Orange}) The contextual embedding network encodes the character labels to produce the contextual semantic matrix for updating vision encoders; (\textcolor{OliveGreen}{Green}) The deep visual representation of characters; (\textcolor{Plum}{Purple}) The classification projector aims to output the predicted labels of input character images.}
\label{fig:3}
\end{figure*}

In the field of language representation, it is widely acknowledged that the semantics of a word are not defined in isolation, but emerge from its contextual usage. This observation has driven the development of contextual language models such as Word2Vec~\cite{ma2015using}, GloVe~\cite{pennington2014glove}, and BERT~\cite{kenton2019bert}, which embed linguistic units into continuous semantic spaces by leveraging co-occurrence or contextual dependencies. In such spaces, semantically related concepts tend to be embedded closer, while unrelated concepts are positioned farther apart. As illustrated in the yellow region of Fig.~\ref{fig:2}, each node represents the contextual semantic embedding, where characters with similar meanings (e.g., \begin{CJK}{UTF8}{gbsn}``圃 (orchard)''\end{CJK} and \begin{CJK}{UTF8}{gbsn}``园 (garden)''\end{CJK}) exhibit higher similarity than semantically distant pairs (e.g., \begin{CJK}{UTF8}{gbsn}``圃 (orchard)''\end{CJK} and \begin{CJK}{UTF8}{gbsn}``帝 (emperor)''\end{CJK}). Importantly, these distances are not binary, but rather reflect a continuum of semantic relevance.

Logographic characters exhibit a unique duality: they convey semantics not only through linguistic context, but also through visual structure. Each character glyph is deliberately designed to encode semantic cues via its visual components, giving rise to what is commonly referred to as \emph{visual semantics}~\cite{diao2023toward}. Characters that express related concepts often share similar radicals or structural components, resulting in higher visual similarity, as shown in the blue region of Fig.~\ref{fig:2}. Moreover, we observe a strong correspondence between contextual and visual semantic space. For instance, the characters of the closely associated concepts orchard and garden are embedded closely in both spaces. This consistency suggests that contextual semantics can provide meaningful guidance for visual representation learning, particularly when visual similarity alone is insufficient to characterize semantic relationships.

\subsection{Motivation and Challenges}

Existing contrastive learning-based pretraining methods for character images typically rely on constructing binary positive and negative sample pairs. The learning objective is to pull positive pairs closer while pushing negative pairs farther apart, thereby shaping the representation space. Given two character images $I_1$ and $I_2$, such objectives can generally be formulated as:
\begin{equation}
T(W) = \sum^P_{k=1} T\big(W,(Y,I_1,I_2)^k\big),
\label{equ:3.1}
\end{equation}
where $W$ denotes the network parameters, $k$ indexes the constructed sample pairs, and $Y$ is a binary indicator defined as:
\begin{equation}
Y =
\begin{cases}
0, & \text{if } I_1 \text{ and } I_2 \text{ belong to the same category}, \\
1, & \text{otherwise}.
\end{cases}
\label{equ:3.2}
\end{equation}

This binary formulation implicitly assumes that all negative pairs are equally dissimilar and should be separated by a fixed margin. As a result, visually or semantically hard negative pairs (e.g., visually similar characters with different meanings) and trivial negative pairs (e.g., visually and semantically unrelated characters) are treated identically during optimization. Such a rigid objective ignores the continuous nature of semantic similarity and fails to reflect the varying informational value of different sample pairs.

This limitation becomes particularly pronounced in long-tailed and imbalanced character datasets, where the number of trivial negative pairs grows disproportionately. During training, these easy pairs dominate the learning signal, while informative but scarce hard pairs are under-emphasized. Consequently, the learned representations tend to be biased toward frequent categories and fail to capture fine-grained semantic structures for low-frequency characters.

Motivated by the observed consistency between contextual semantics and visual semantics, we argue that an effective learning objective should account for the degree of semantic relevance between samples, rather than enforcing a binary decision. To this end, we introduce a soft semantic supervision mechanism derived from contextual semantic space, which provides continuous similarity targets for character pairs:
\begin{equation}
T(W) = \sum^P_{k=1} T\big(W,(CS(I_1,I_2), I_1, I_2)^k\big),
\label{equ:3.3}
\end{equation}
where $CS(I_1,I_2) \in [0,1]$ denotes the contextual semantic relation between the two characters.

By replacing binary labels with soft semantic supervision, the proposed formulation enables the learning objective to distinguish between hard and trivial sample pairs, thereby allowing a more efficient allocation of learning capacity under imbalanced data distributions. Importantly, the introduced contextual semantics are not used as absolute regression targets for visual representations. Due to the inherent modality gap between language and vision, directly aligning visual features with contextual embeddings is impractical and potentially misleading. Instead, we exploit pre-trained language models to extract relative semantic relations between characters by measuring similarities in the contextual semantic space. These relations provide continuous, relational supervision signals that guide the learning objective of visual representations without enforcing direct embedding alignment. In this way, contextual semantics serve as a learnable guidance mechanism rather than an absolute target, naturally facilitating semantic-aware contrastive learning and laying the foundation for the contextual distribution alignment strategy introduced in the following section.

\section{The Proposed Method}
\label{sec:Method}






In this section, we introduce our methodology, with a primary focus on the pre-training strategy designed to enhance the semantic representation of logographic characters. Our method guides the learning of deep visual representations under the constraints of soft labels by leveraging both visual and contextual semantics in the latent space. We reformulate contextual semantic supervision as a soft contrastive learning objective, where contextual similarities define soft positive and negative relations for each anchor. Figure~\ref{fig:3} illustrates the proposed pre-training framework and its integration with downstream character recognition.

Given a character image $I \in \mathbb{R}^{H\times W\times 3}$, its label is denoted as $L$.
We aims to enhance the deep visual representation of logographic character images by aligning the corresponding visual semantics and contextual semantics. In particular, we reformulate contextual semantic supervision as a distribution-level contrastive objective, which is more suitable for learning under long-tailed data distributions. The character image and its label undergo two crucial transformations: obtaining visual semantic representation $V_s$, and contextual semantic representation $C_s$. Considering that visual and contextual representations are obtained from sources with different modalities, a direct computation might lead to suboptimal results. Thus, we introduce a semantic calculation to establish the mapping between visual and contextual semantics. The semantic calculation quantifies the consistency of underlying concepts shared between visual and contextual semantics.

\subsection{Visual Semantic Similarity}
To capture the distinctive visual features intrinsic to logographic characters, we exploit a vision encoder to extract visual semantics from character images. 
A neural network-based encoder $VisionEncoder(\cdot)$ is employed to extract representation vectors from augmented character image samples.
Mathematically, it is expressed as:
\begin{equation}
h = VisionEncoder(I),
\label{equ:4.0}
\end{equation}
where $h$ denotes the output after the average pooling layer. 
A projection head $f(\cdot)$, implemented as a multi-layer perceptron (MLP) with one hidden layer, is further applied to map the representations into the embedding space where semantic alignment is performed:
\begin{equation}
V_s = f(h).
\label{equ:4.1}
\end{equation}
Here, $V_s$ represents the visual representation of the input character image $I$, encoding its visual semantics.
Our framework does not impose any restriction on the choice of the vision backbone. 
For simplicity and efficiency, we adopt ResNet~\cite{he2016deep} as the vision encoder in our implementation.

To quantify the visual semantic relations between characters, we compute pairwise cosine similarities between visual embeddings:
\begin{equation}
Sem^{i,j}_V = \frac{V_i \cdot V_j}{\|V_i\| \|V_j\|}, \quad i,j \in [1,B],
\label{equ:4.2}
\end{equation}
where $V_i$ and $V_j$ denote the visual semantic representations of the $i^{th}$ and $j^{th}$ character images within a mini-batch, respectively, and $B$ is the mini-batch size. The resulting visual semantic similarity matrix $Sem_V \in \mathbb{R}^{B \times B}$ is computed independently for each mini-batch and reflects the pairwise visual relations among the sampled character images. Note that cosine similarity is only used to compute relative affinities between samples. It does not directly serve as the optimization target.

\subsection{Contextual Semantic Distribution}
We leverage state-of-the-art language models to extract contextual semantics from character labels and utilize them as semantic supervision for visual representation learning. 
Specifically, given a character label represented in natural language, a contextual embedding network $ContextEncoder(\cdot)$ is applied to obtain its contextual semantic representation:
\begin{equation}
C_s = ContextEncoder(L),
\label{equ:4.3}
\end{equation}
where $L$ denotes the textual label of the character, and $ContextEncoder(\cdot)$ corresponds to a pre-trained language model. 
In this work, we adopt BERT as the contextual encoder, not to directly obtain absolute text embeddings, but to leverage its strong contextual modeling capability to capture relative semantic relations among characters.

Based on contextual embeddings, we first compute pairwise semantic similarity scores using cosine similarity:
\begin{equation}
Sem^{i,j}_C = \frac{C_i \cdot C_j}{\|C_i\| \|C_j\|}, \quad i,j \in [1,B],
\label{equ:4.4}
\end{equation}
where $C_i$ and $C_j$ are the contextual semantic representations of the $i^{th}$ and $j^{th}$ character labels, respectively. 
The resulting matrix $Sem_C \in \mathbb{R}^{B \times B}$ captures raw contextual affinity scores between the corresponding character labels within the mini-batch. However, cosine similarity is inherently bounded and may suffer from non-isometric scaling, making it difficult to adequately distinguish informative semantic pairs from a large number of trivial ones, especially under long-tailed distributions.

Rather than directly regressing these similarity scores, which may suffer from scale inconsistency and non-isometric embedding distributions, we interpret the contextual similarities as defining a conditional probability distribution over semantic neighbors. 
For each anchor character $i$, the contextual semantic distribution is obtained by applying a temperature-scaled softmax function:
\begin{equation}
P_C(j|i) = 
\frac{\exp\left(Sem^{i,j}_C / \tau_C\right)}
{\sum_{k \neq i} \exp\left(Sem^{i,k}_C / \tau_C\right)},
\label{equ:4.5}
\end{equation}
where $\tau_C$ is a temperature parameter controlling the sharpness of the semantic distribution.
The distribution $P_C(\cdot|i)$ characterizes the relative semantic relevance of other characters with respect to the anchor character $i$, and serves as a soft semantic supervision signal for guiding visual representation learning. By utilizing temperature-scaled distribution, this formulation emphasizes relative semantic ordering and effectively amplifies informative, underrepresented semantic relations while suppressing trivial pairs.

\subsection{Contextual Distribution Alignment Loss}
As discussed in the previous sections, directly regressing pairwise semantic similarity scores may be sensitive to scale variations and does not explicitly preserve the relative semantic structure among characters. Instead, we formulate contextual semantic supervision at the distribution level and align the semantic neighborhood distributions between contextual and visual modalities.

Given the contextual semantic similarity matrix $Sem_C$ and the visual semantic similarity matrix $Sem_V$, we construct conditional semantic distributions for both modalities.
For each anchor character $i$, the contextual semantic distribution is defined as Function \ref{equ:4.5}. Similarly, the visual semantic distribution is obtained as:
\begin{equation}
P_V(j|i) = 
\frac{\exp\left(Sem^{i,j}_V / \tau_V\right)}
{\sum_{k \neq i} \exp\left(Sem^{i,k}_V / \tau_V\right)},
\label{equ:4.7}
\end{equation}
where $\tau_V$ controls the sharpness of the visual semantic distribution.

We formulate contextual semantic supervision as a distribution-level contrastive objective by aligning the visual semantic distribution $P_V(\cdot|i)$ with the contextual semantic distribution $P_C(\cdot|i)$. Specifically, we minimize the cross-entropy between the two distributions:
\begin{equation}
\mathcal{L}_{CS}^{\text{Vanilla}}
=
-\frac{1}{B}
\sum_{i=1}^{B}
\sum_{j\neq i}
P_C(j|i)\,
\log P_V(j|i),
\label{equ:cs_kl}
\end{equation}
This formulation is equivalent to minimizing the Kullback--Leibler divergence between the contextual and visual semantic distributions, and can be interpreted as a soft-label contrastive learning objective that preserves relative semantic ranking structures rather than absolute similarity values. In this formulation, the soft labels are derived from cosine-based contextual semantic affinities and integrated into the contrastive objective.

We adopt cosine similarity to measure pairwise semantic affinity, as it provides a scale-invariant and direction-aware metric that is widely used in embedding learning. However, cosine similarity alone only captures point-wise relationships and does not explicitly model the relative semantic structure among multiple samples. To address this limitation, we further normalize cosine similarities using a temperature-scaled softmax to construct a semantic neighborhood distribution for each anchor. The resulting distributions are then aligned using Kullback--Leibler divergence, which enables structure-level supervision by preserving relative semantic ordering rather than absolute similarity magnitudes. This design is particularly suitable for long-tailed character distributions, where learning robust semantic neighborhood structures is more critical than fitting individual similarity values.

\noindent \textbf{Difficulty-aware Focal Reweighting.}
Although distribution-level semantic alignment effectively captures contextual relationships, long-tailed character datasets typically contain a large number of semantically trivial pairs that are easy to align. Treating all semantic pairs equally may lead to inefficient optimization, as easy pairs dominate the loss while informative hard pairs are under-emphasized. To address this issue, we introduce an optional difficulty-aware reweighting strategy inspired by focal loss, which adaptively modulates the contribution of each semantic pair based on the current prediction confidence of the visual model.

Concretely, we define a focal weighting term based on the visual semantic distribution $P_V(j|i)$, and reweight the contextual semantic loss as follows:
\begin{equation}
\mathcal{L}_{CS}^{\text{FR}}
=
-\frac{1}{B}
\sum_{i=1}^{B}
\sum_{j\neq i}
P_C(j|i)\,
\big(1 - P_V(j|i)\big)^{\gamma}\,
\log P_V(j|i),
\end{equation}
where $\gamma$ is the focusing parameter. The focal term down-weights semantic pairs that are already well predicted by the visual model, and emphasizes harder pairs that remain misaligned. Importantly, this reweighting does not alter the semantic supervision provided by the contextual encoder, but only adjusts the relative importance of semantic pairs according to their learning difficulty. In all experiments, we set $\gamma = 1$ and observe stable performance across datasets.

\subsection{Classification Loss}

In addition to contextual semantic supervision, we introduce a classification projector $\Phi(\cdot)$ to perform character-level recognition and provide task-specific guidance during training.
The classification head maps the learned visual representations to the label space and is optimized using the standard cross-entropy loss:
\begin{equation}
\mathcal{L}_{Cls} = -\sum_{m=1}^{M} \Phi(y_m) \cdot \log\big(\hat{\Phi(p_m)}\big),
\label{equ:4.6}
\end{equation}
where $M$ denotes the number of character categories in the dataset, $\Phi(y_m)$ represents the one-hot encoded ground-truth label, and $\hat{\Phi(p_m)}$ is the predicted probability for the $m$-th category. The classification loss serves as a complementary objective that anchors the learned representations to the target recognition task, while the contextual semantic loss focuses on preserving relative semantic structures in the embedding space.
During training, the two objectives are jointly optimized, resulting in the overall loss: $\mathcal{L} {=} \mathcal{L}_{CS} {+} \mathcal{L}_{Cls}$.


\section{Experiments}
\label{sec:Experiments}

\begin{table*}[t]
\setlength{\abovecaptionskip}{5pt}%
\setlength{\belowcaptionskip}{0pt}%
\centering
\caption{Quantitative comparisons of logographic character recognition accuracy (\%) with state-of-the-art image representation and contrastive learning-based pre-training methods on four long-tail datasets. R18 and R32 denote ResNet18 and ResNet32 backbones, respectively.}
\label{tab:1.1}
\resizebox{1\linewidth}{!}{%
\begin{tabular}{@{}lccccccccccccccccc@{}}
\toprule
\multicolumn{1}{c}{\multirow{3}{*}{Methods}}                  &    & \multicolumn{6}{c}{HWDB1.1}                                                                         & \multicolumn{6}{c}{ICDAR2013}                                                                       & \multicolumn{2}{c}{\multirow{2}{*}{CTW}} & \multicolumn{2}{c}{\multirow{2}{*}{K-Kanji}} \\ \cmidrule(l){2-2}  \cmidrule(l){3-8} \cmidrule(l){9-14} 
& IF & \multicolumn{2}{c}{100}      & \multicolumn{2}{c}{50}       & \multicolumn{2}{c}{10}       & \multicolumn{2}{c}{100}      & \multicolumn{2}{c}{50}       & \multicolumn{2}{c}{10}       & \multicolumn{2}{c}{}                     & \multicolumn{2}{c}{}                         \\ \cmidrule(l){2-2} \cmidrule(l){3-4} \cmidrule(l){5-6} \cmidrule(l){7-8} \cmidrule(l){9-10} \cmidrule(l){11-12} \cmidrule(l){13-14} \cmidrule(l){15-16}  \cmidrule(l){17-18}  
 & BB & R18            & R32            & R18            & R32            & R18            & R32            & R18            & R32            & R18            & R32            & R18            & R32            & R18            & R32            & R18            & R32            \\   \midrule
SimClr   \cite{chen2020simple}     &    & 56.68          & 59.67          & 68.85          & 70.28          & 82.42          & 83.89          & 58.65          & 61.04          & 68.41          & 69.34          & 83.15          & 85.20          & 69.37          & 72.06          & 82.81          & 83.34          \\
MOCO \cite{he2020momentum}         &    & 58.96          & 62.01          & 74.26          & 76.40          & 84.91          & 85.88          & 62.13          & 64.53          & 74.68          & 76.74          & 85.23          & 87.06          & 70.63          & 74.82          & 84.82          & 85.13          \\
BYOL    \cite{grill2020bootstrap}   &    & 57.32          & 59.86          & 76.56          & 78.90          & 85.63          & 88.04          & 61.32          & 62.83          & 76.82          & 79.06          & 85.96          & 88.27          & 68.96          & 73.12          & 83.28          & 84.75          \\
SwAV   \cite{caron2020unsupervised}                &    & 62.56          & 64.23          & 75.72          & 78.65          & 83.65          & 85.43          & 63.57          & 64.65          & 76.23          & 78.92          & 84.66          & 86.12          & 72.47          & 75.14          & 86.63          & 87.32          \\
SimSiam   \cite{chen2021exploring} &    & 63.07          & 65.84          & 77.19          & 79.84          & 83.78          & 85.65          & 64.81          & 66.40          & 77.48          & 79.80          & 84.34          & 85.87          & 73.49          & 77.21          & 86.16          & 86.64          \\      
VL-Bert \cite{su2019vl}      &    &  76.73         & 78.65          & 81.37          & 82.12           & 84.84          & 86.30          & 76.85          & 78.96         &  81.68       & 83.29          & 84.95          & 86.43          &  79.85          &  83.94       & 90.43         & 92.72              \\
CLIP \cite{radford2021learning}    &    &  79.45          &  81.22         & 83.54          & 84.59          & 85.95          & 88.86          & 81.86          & 83.73         & 84.79        & 86.04          & 89.35          &  90.46         &  83.20         &  85.82       & 93.35         & 94.38              \\ \midrule
CMO \cite{park2022majority}        &    & 65.85          & 68.16          & 79.68          & 81.03          & 84.82          & 85.94          & 66.88          & 68.74          & 78.94          & 80.63          & 84.13          & 85.69          & 75.06          & 78.45          & 88.83          & 89.41          \\
OTmix \cite{gao2024enhancing}      &    & 68.72          & 70.04          & 80.32          & 82.78          & 85.91          & 87.24          & 69.40          & 71.23          & 80.65          & 83.08          & 86.32          & 88.24          & 76.67          & 79.72          & 89.62          & 90.03          \\
FCC \cite{li2023fcc}               &    & 72.65          & 74.36          & 82.23          & 84.62          & 87.76          & 89.13          & 72.23          & 71.21          & 81.71          & 83.69          & 86.65          & 88.82          & 77.28          & 81.89          & 88.04          & 89.13          \\
COCL \cite{miao2024out}            &    & 78.44          & 80.02          & 85.36          & 87.51          & 88.38          & 90.89          & 76.13          & 79.12          & 84.04          & 86.22          & 86.65          & 89.32          & 75.54          & 79.43          & 90.82          & 91.68          \\
NCL++ \cite{tan2024ncl++}          &    & 75.32          & 78.48          & 83.46          & 84.91          & 86.94          & 88.67          & 78.84          & 79.69          & 85.19          & 87.61          & 88.21          & 90.43          & 78.65          & 82.86          & 92.03          & 94.21          \\
ProCo   \cite{du2024probabilistic} &    & 76.88          & 79.14          & 84.59          & 86.47          & 88.23          & 89.72          & 77.46          & 79.89          & 84.82          & 86.83          & 89.14          & 90.58          & 79.74          & 83.83          & 91.86          & 92.04          \\    \midrule
Ours$_{Vanilla}$                                             &    
& 82.63 & 83.97 & 88.35 & 90.04 & 91.68 & 92.53 
& 83.89 & 86.02 & 88.82 & 89.67 & 93.26 & 92.57
& 85.33 & 88.12 & 92.89 & 94.75         \\
Ours$_{FR}$                                                &    
& \textbf{83.46} & \textbf{85.28} & \textbf{89.80} & \textbf{90.92} & \textbf{92.29} & \textbf{93.80} 
& \textbf{84.37} & \textbf{86.61} & \textbf{89.84} & \textbf{91.03} & \textbf{93.70} & \textbf{93.09} 
& \textbf{86.97} & \textbf{88.60} & \textbf{94.33} & \textbf{95.30}          \\ \bottomrule
\end{tabular}}
\end{table*}

\subsection{Experimental Setup}
\noindent\textbf{Datasets.}
To comprehensively evaluate the visual representation of logographic characters extracted by our proposed pre-training method, we strategically selected four datasets to cover a broad range of sources, including HWDB1.1 \cite{liu2013online}, ICDAR2013 \cite{yin2013icdar}, scene character dataset CTW \cite{yuan2019large}, and Japanese Historical Character image dataset Kuzushiji-Kanji (K-Kanji) \cite{saini2019japanese}.
Specifically, the HWDB1.1 and ICDAR2013 focus on handwritten Chinese characters. To assess performance on imbalanced datasets with varying degrees, we created three distinct long-tail training sets for these two datasets by altering the imbalance factor\footnote{Defined as the number of training samples in the largest class divided by the number in the smallest class.} $IF \in \{10, 50, 100\}$. 
For instance, we use HWDB-LT100 to denote a long-tailed version of HWDB1.1, where the number of samples per class follows a power-law distribution, resulting in a significant imbalance between head and tail classes, with an IF of 100.
We also conducted experiments on datasets with imbalanced distributions in the real world. The CTW was collected in natural settings to incorporate real-world scenes, such as road signs and billboards. The CTW consists of a total of 3,650 classes, with 1 to 13,900 samples per class, an obvious class imbalance.
The K-Kanji dataset is a highly imbalanced dataset extracted from 44 Japanese historical books, featuring 3,832 unique Kanji character categories and 140,426 samples in Japanese. The dataset includes both common and rare characters, with category distributions ranging from 1,766 instances to just one example per category. This dataset, derived from formal historical records, captures authentic character distributions.

\noindent\textbf{Implementation Details.}
We implemented all the networks and training procedures in PyTorch, and performed all experiments on eight NVIDIA A100 GPUs. The resolution of the input image is $96 {\times} 96$. In our approach, we exploit basic data augmentation techniques, including translation, rotation, scaling, background removal, and background transformation. In the pre-training stage, we set the initial learning rate at $3e-5$.
We employ the OneCycleLR strategy for learning rate adjustment, with a maximum learning rate $lr_{max}=0.0001$. Following common practice in contrastive learning, we set the temperature parameter $\tau$ to 0.07.

\subsection{Experimental Results}
\noindent\textbf{Comparison with pre-training methods.}
In this experiment, we compare our proposed method against several well-established pre-training strategies, including contrastive learning-based pretraining methods, e.g., SimClr \cite{chen2020simple}, MOCO \cite{he2020momentum}, BYOL \cite{grill2020bootstrap}, SwAV \cite{caron2020unsupervised}, and SimSiam \cite{chen2021exploring}, and multimodal pretraining models, e.g., VL-BERT \cite{su2019vl} and CLIP \cite{radford2021learning} (using the same training setup from \cite{yu2023chinese}). 
To replicate the experimental setups of these methods and facilitate comparison, we employed ResNet18 (R18) and ResNet32 (R32) as the baseline models for all aforementioned methods. 

We evaluated these methods on four long-tail datasets under three different imbalance ratios (IF = 100, 50,10, respectively), as shown in Table~\ref{tab:1.1}. 
%
Overall, our method consistently outperforms all competing baselines across different datasets, imbalance levels, and backbone architectures.
Compared to contrastive learning-based pre-training methods, our approach achieves substantial performance gains, with particularly notable improvements under more severe long-tail settings. To further analyze the contribution of each component, we report two variants of our method. \textit{Ours$_{Vanilla}$} corresponds to the proposed context-aware pre-training framework based on distribution-level semantic alignment, where soft semantic labels derived from language models are integrated into a contrastive learning objective. This variant already yields significant improvements over all baseline methods, demonstrating the effectiveness of modeling relative semantic relations for long-tailed character recognition. Building upon this formulation, \textit{Ours$_{FR}$} additionally incorporates a difficulty-aware focal reweighting strategy to adaptively emphasize informative and hard semantic relations during training. As shown in Table~\ref{tab:1.1}, this focal reweighting consistently leads to further performance improvements over \textit{Ours$_{Vanilla}$}, especially under more imbalanced settings (IF = 100 and 50) and for smaller backbones (R18). This observation indicates that selectively amplifying hard semantic pairs is particularly beneficial for mitigating the dominance of trivial pairs in long-tailed distributions.
%


\begin{table*}[t]
\centering
\setlength{\abovecaptionskip}{5pt}%
\setlength{\belowcaptionskip}{0pt}%
\caption{The recognition accuracy (\%) of logographic characters was quantitatively compared across three logographic character datasets, evaluating the SOTA OCR methods with and without our proposed pretraining method. ``Impro.(\%)"  is the improvement in performance. }
\label{tab:3}
\resizebox{0.95\linewidth}{!}{%
\begin{tabular}{l|ccc|ccc|ccc}
\toprule
\multicolumn{1}{c|}{Datasets}                              & \multicolumn{3}{c|}{HWDB1.1} & \multicolumn{3}{c|}{ICDAR2013} & \multicolumn{3}{c}{CTW} \\ \midrule
\multicolumn{1}{c|}{Pre-training}                          & Without               & With ours             & Impro.(\%)          & Without                & With ours              & Impro.(\%)            & Without             & With ours            & Impro.(\%)          \\ \midrule
AlexNet   \cite{krizhevsky2012imagenet}   & 88.74            & 90.35            & + 1.61           & 89.99             & 90.86            & + 0.87            & 76.49          & 78.16          &  + 1.67          \\
VGG16 \cite{simonyan2014very}             & 89.67            & 91.23            &  + 1.56           & 90.68             & 91.45            &  + 0.77            & 79.38          & 80.46          &  + 1.08          \\
HCCR-GoogLeNet   \cite{zhong2015high}     & 94.85            & 95.13            &  + 0.28           & 96.26             & 97.12            &  + 0.86            & 82.28          & 82.89          &  + 0.61          \\
ResNet \cite{he2016deep}                  & 90.98            & 92.28            &  + 1.30           & 92.18             & 92.74            &  + 0.56            & 79.46          & 81.08          &  + 1.62          \\
DropSample-DCNN \cite{yang2016dropsample}  & 96.57            & 96.92            &  + 0.35           & 97.23             & 97.56            &  + 0.33            & 82.37          & 83.23          &  + 0.86          \\
DirectMap \cite{zhang2017online}           & 96.25            & 96.95            &  + 0.70           & 97.37             & 97.68            &  + 0.31            & 84.23          & 85.04          &  + 0.81          \\
M-RBC + IR \cite{yang2017improving}      & 96.14            & 96.57            &  + 0.43           & 97.37             & 97.73            &  + 0.36            & 83.65          & 84.23          &  + 0.58          \\
DenseNet   \cite{huang2017densely}        & 94.32            & 95.13            &  + 0.81           & 95.90             & 96.81            &  + 0.91            & 79.45          & 81.52          &  + 2.07          \\
ViT   \cite{dosovitskiy2020image}         & 95.13            & 95.63            &  + 0.50           & 96.26             & 97.42            &  + 1.16            & 79.80          & 81.65          &  + 1.85          \\
Stroke-to-Character   \cite{chen2021zero} & 93.97            & 94.24            &  + 0.27           & 96.28             & 96.55            &  + 0.27            & 85.29          & 85.78          &  + 0.49          \\
CUE  \cite{luo2023self}                    & 95.23            & 96.45            &  + 1.22           & 96.96             & 97.73            &  + 0.77            & 83.81          & 84.23          &  + 0.42          \\
RZCR  \cite{diao2023rzcr}               & 95.27            & 96.13            &  + 0.86           & 96.35             & 97.53            &  + 1.18            & 86.64          & 87.16          &  + 0.52          \\
UCR  \cite{li2025ucr}                      & 96.92            & 97.37            &  + 0.45           & 96.55             & 97.59            &  + 1.04            & 88.29          & 89.25          &  + 0.96          \\ \bottomrule
\end{tabular}}
\end{table*}

\noindent\textbf{Comparison with Imbalanced Image Classification Methods.}
We further compare our approach with several recent state-of-the-art representation learning methods specifically designed for long-tailed visual recognition, including CMO \cite{park2022majority}, OTmix \cite{gao2024enhancing}, FCC \cite{li2023fcc}, COCL \cite{miao2024out}, NCL++ \cite{tan2024ncl++}, and ProCo \cite{du2024probabilistic}. These methods primarily address class imbalance through classification-level reweighting or feature regularization strategies. As reported in Table~\ref{tab:1.1}, our method consistently achieves the best performance across all datasets and imbalance settings using both R18 and R32 backbones. We observe that long-tail classification methods such as NCL++ and ProCo outperform generic contrastive pre-training baselines, indicating the effectiveness of imbalance-aware optimization at the classifier level. However, in the context of logographic character recognition, these methods remain limited by their reliance on visual sample distributions alone and lack explicit mechanisms to model semantic relationships among character categories. As a result, their performance degrades when visual samples are extremely sparse or unevenly distributed.
Overall, these results demonstrate that integrating contextual semantics with representation-level long-tail modeling offers superior robustness and scalability for real-world logographic character recognition.

\noindent \textbf{Comparison of OCR Models With and Without Pretraining.}
To evaluate the effectiveness of our proposed pretraining method for ideographic character recognition, we conducted a comprehensive comparison across HWDB1.1, ICDAR2013, and CTW. As shown in Table~\ref{tab:3}, we benchmarked a range of SOTA OCR models, including both CNN-based (e.g., AlexNet, VGG16, ResNet, DenseNet) and transformer-based architectures (e.g., ViT, CUE, UCR), with and without the incorporation of our pretraining strategy (vanilla version).
Across all models and datasets, the integration of our method consistently improved recognition accuracy. For instance, AlexNet achieved a notable improvement of 1.61\%, 0.87\%, and 1.67\% on HWDB1.1, ICDAR2013, and CTW respectively. Similarly, advanced models such as UCR and RZCR also benefited from our method, with performance gains up to 1.18\% on ICDAR2013 and 0.96\% on CTW. These results demonstrate the generalizability and effectiveness of our approach across various OCR backbones and datasets.
The improvements are particularly significant on challenging datasets like CTW, where models such as DenseNet and ViT obtained gains of 2.07\% and 1.85\%, respectively. These findings suggest that our pretraining method effectively enhances the model’s ability to capture the structural semantics. 

\begin{table}[t]
\centering
\setlength{\abovecaptionskip}{5pt}%
\setlength{\belowcaptionskip}{10pt}%
\caption{Ablation study for the choice of loss functions, The best results are highlighted in \textbf{bold}.}
\label{tab:4}
\resizebox{1\linewidth}{!}{%
\begin{tabular}{@{}cc|cccc@{}}
\toprule
classification loss & contextual loss        & HWDB-LT100 & ICDAR-LT100 & CTW &  K-Kanji  \\ \midrule
 \checkmark    &  & 74.38          & 72.67           & 67.21          & 80.24     \\
  & \checkmark         & 82.42          & 83.36           & 84.74          & 92.40      \\
 \checkmark   & \checkmark      & \textbf{82.63} & \textbf{83.89}  & \textbf{85.33} & \textbf{92.89}   \\ 
 \bottomrule
\end{tabular}}
\end{table}

\begin{table}[t]
\centering
\setlength{\abovecaptionskip}{5pt}%
\setlength{\belowcaptionskip}{10pt}%
\caption{Ablation study on different distance metrics for contextual semantic alignment. The best results are highlighted in \textbf{bold}.}
\label{tab:5}
\resizebox{1\linewidth}{!}{%
\begin{tabular}{@{}c|cccc@{}}
\toprule
Contextual loss & HWDB-LT100 & ICDAR-LT100 & CTW & K-Kanji \\ \midrule
point-wise (MSE)  & 79.46 & 78.92 & 80.50 & 90.02 \\
distribution-level & \textbf{82.63} & \textbf{83.89} & \textbf{85.33} & \textbf{92.89} \\
\bottomrule
\end{tabular}}
\end{table}

\subsection{Ablation Study}

\noindent \textbf{Effect of loss functions.} 
Our pretraining method employs two types of loss function to achieve optimal character recognition performance. To assess the impact of using different losses on our proposed method, we conduct an ablation study by evaluating the effects of using classification loss, contextual loss, and their combination. 
The results of this experiment are shown in Table~\ref{tab:4}, where the marks indicate the use of each loss. We observed that when only classification loss is used, performance is relatively limited in all datasets, with accuracy scores of 74.38\% in HWDB-LT100 and 72.67\% in ICDAR-LT100. In contrast, adopting only contextual loss significantly improves performance, reaching 82.42\% and 83.36\% in the same datasets, respectively. This demonstrates the effectiveness of leveraging semantic and structural context in enhancing character representation.
Additionally, we found that applying both losses together yielded the best results. This is because contextual loss helps to inherently align more closely with the two semantic representations of text: visual and contextual semantics. This alignment led to better character representations, thus improving recognition performance. This experiment demonstrates the effectiveness of the proposed contextual loss.
These results highlight the importance of joint optimization for robust logographic character recognition in long-tail scenarios.

\noindent \textbf{Effect of distance metrics for semantic alignment.}
To further investigate how different alignment objectives affect representation learning, we conduct an ablation study by comparing distance-based and distribution-based semantic alignment strategies.
Specifically, we replace the proposed distribution-level contrastive loss with a mean squared error (MSE) loss that directly regresses pairwise semantic similarity scores between the contextual and visual modalities.
\begin{equation}
\mathcal{L}_{CS}^{\text{MSE}}
=
\frac{1}{B}
\sum_{i=1}^{B}
\sum_{j \neq i}
\left(
Sem_C(i,j) - Sem_V(i,j)
\right)^2,
\label{equ:mse_cs}
\end{equation}
The MSE-based objective enforces point-wise consistency between similarity values and can be interpreted as learning absolute distances in the embedding space. In contrast, the proposed KL-based (equation \ref{equ:cs_kl}) objective aligns semantic neighborhood distributions, emphasizing relative semantic ordering rather than exact similarity magnitudes. This distinction is particularly important in long-tailed settings, where embedding spaces are often non-isometric and dominated by trivial pairs. As shown in Table~\ref{tab:5}, the MSE-based alignment consistently underperforms the distribution-level contrastive objective (vanilla version) across all evaluated datasets. This result indicates that directly regressing semantic distances is less effective than learning relational structures among samples. Overall, this ablation confirms that modeling semantic supervision as a distribution-level contrastive objective is more suitable for capturing meaningful spatial relationships and leads to more robust representations for long-tailed character recognition.

\section{Conclusion}
\label{sec:Conclusion}
This paper introduces a contrastive pretraining method, aimed at addressing the task of imbalanced logographic character recognition in real-world scenarios. We begin by discussing how current character recognition approaches are challenged by imbalanced character datasets in the real world. We then introduce the correspondence between visual semantics and contextual semantics in logographic character languages and how it can influence the ability of deep neural networks to capture features of character images. Based on these observations, we propose a novel context-aware pretraining strategy designed to enhance the performance and learning efficiency of imbalanced character classes in their deep representations. This strategy seeks to improve visual representations by leveraging the contextual semantics of each character extracted from corresponding language models as soft labels. Finally, we demonstrate the superiority of our method over other approaches through multiple experiments, particularly on logographic character datasets with imbalanced data distribution.




\bibliographystyle{ACM-Reference-Format}
\balance
\bibliography{sample-base}










\end{document}